\pdfoutput=1

\documentclass[11pt]{article}

\usepackage[unicode]{hyperref}

\usepackage[preprint]{acl}

\usepackage{times}
\usepackage{latexsym}

\usepackage[T1]{fontenc}

\usepackage[utf8]{inputenc}

\usepackage{microtype}

\usepackage{inconsolata}

\usepackage{graphicx}

\usepackage{multirow}
\usepackage{amsmath}
\usepackage{amssymb}
\usepackage{here}
\usepackage{float}
\usepackage{colortbl}
\usepackage{blindtext}
\usepackage{multicol}
\usepackage{float}
\usepackage{enumitem}
\usepackage{tabularx}

\newcommand{\tabH}{\rule{0pt}{2.3ex}}

\newcommand{\bhline}{\noalign{\hrule height 1.2pt}}
\newcommand{\footlink}[1]{\footnote{\url{#1}}}

\newcommand{\hf}[1]{\href{https://huggingface.co/#1}{#1}}

\newcommand{\comment}[1]{\textcolor{black}{#1}}

\title{
Redundancy, Isotropy, and Intrinsic Dimensionality\\of Prompt-based Text Embeddings
}

\author{
Hayato Tsukagoshi \hspace{4ex} Ryohei Sasano \\
Graduate School of Informatics, Nagoya University \\
\texttt{tsukagoshi.hayato.r2@s.mail.nagoya-u.ac.jp} \hspace{2ex} \texttt{sasano@i.nagoya-u.ac.jp}\\
}

\begin{document}
\maketitle

\begin{abstract}
Prompt-based text embedding models, which generate task-specific embeddings upon receiving tailored prompts, have recently demonstrated remarkable performance.
However, their resulting embeddings often have thousands of dimensions, leading to high storage costs and increased computational costs of embedding-based operations.
In this paper, we investigate how post-hoc dimensionality reduction applied to the embeddings affects the performance of various tasks that leverage these embeddings, specifically classification, clustering, retrieval, and semantic textual similarity (STS) tasks.
Our experiments show that even a naive dimensionality reduction, which keeps only the first 25\% of the dimensions of the embeddings, results in a very slight performance degradation, indicating that these embeddings are highly redundant.
Notably, for classification and clustering, even when embeddings are reduced to less than 0.5\% of the original dimensionality the performance degradation is very small.
To quantitatively analyze this redundancy, we perform an analysis based on the intrinsic dimensionality and isotropy of the embeddings.
Our analysis reveals that embeddings for classification and clustering, which are considered to have very high dimensional redundancy, exhibit lower intrinsic dimensionality and less isotropy compared with those for retrieval and STS.
\end{abstract}

\begin{figure}[t]
\includegraphics[width=\linewidth]{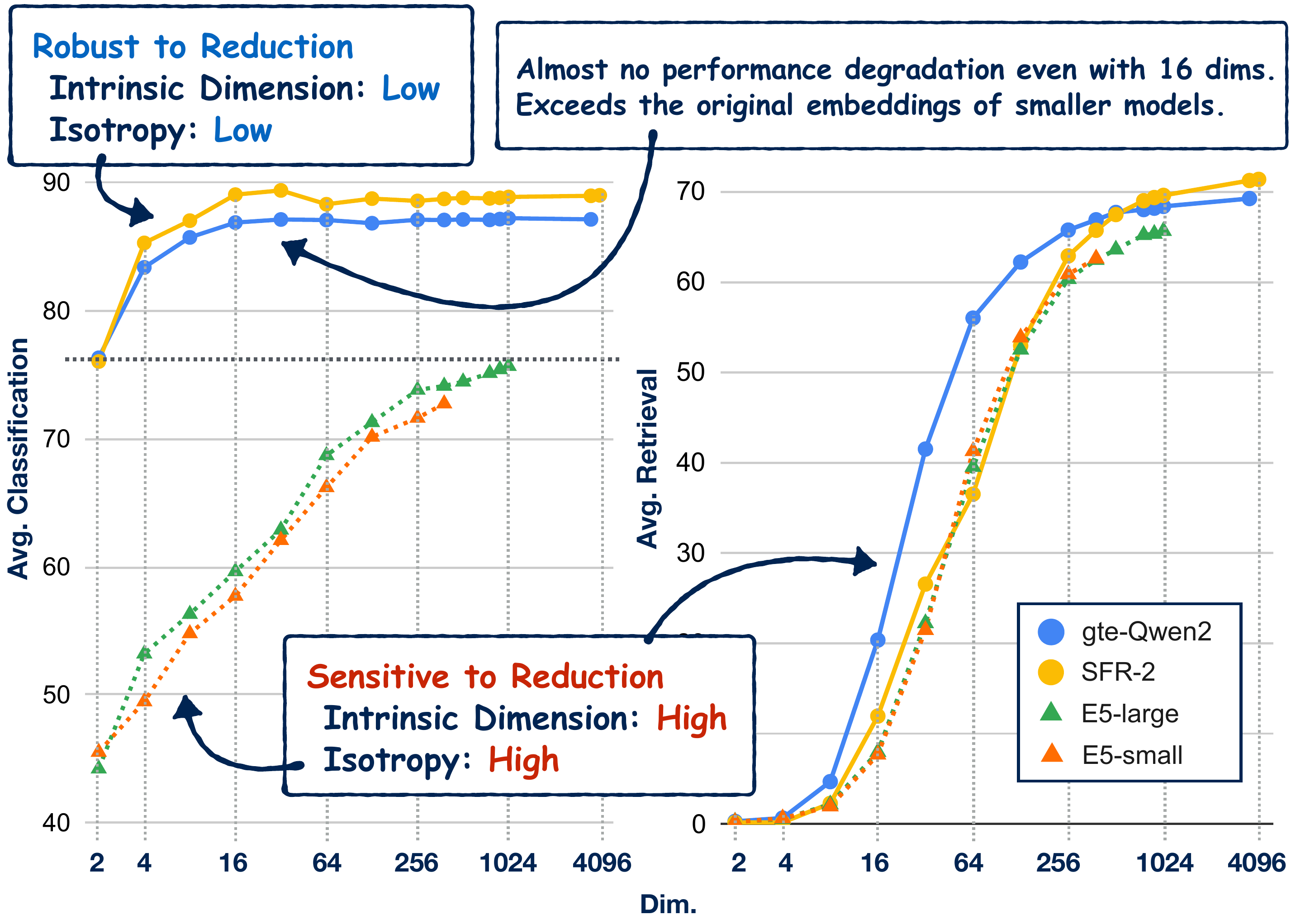}
\caption{
The performance on classification and retrieval tasks under dimensionality reduction.
The models used in the experiments are described in Section~\ref{sec:reduction}.
} 
\label{fig:overview}
\end{figure}

\section{Introduction}
Text embeddings are a foundational component of many natural language processing (NLP) applications, including document retrieval, retrieval-augmented generation (RAG), and text clustering.
Recent advances in large language models (LLMs) renewed interest in text representation learning, owing to their strong language understanding and generalization capabilities~\cite{SGPT,GTR,ST5,mST5,EchoEmbedding,PromptEOL}.
Among these, prompt-based text embedding models, which produce task-specific embeddings by incorporating natural language instructions or task descriptions, have demonstrated remarkable performance on various tasks~\cite{InstructOR,TART,NV-Embed,Gecko,E5,GTE,BGE}.
However, prompt-based models typically generate embeddings with thousands of dimensions, leading to high storage costs and increased computational costs of embedding-based operations.
For instance, E5-mistral~\cite{Mistral-E5}, a model obtained by fine-tuning Mistral-7B~\cite{Mistral}, produces 4096-dimensional embeddings.
Reducing the dimensionality of LLM-based text embeddings through post-processing could thus offer substantial practical benefits.

In this work, we show that prompt-based text embedding models can maintain surprisingly strong performance even when their dimensionality is substantially reduced.
Figure~\ref{fig:overview} summarizes our findings.
Even a naive dimensionality reduction that keeps only the first 25\% of the embedding dimensions results in almost no performance degradation across a range of tasks, indicating that these embeddings are highly redundant.
Notably, for classification task, reducing embeddings to less than 0.5\% of the original dimensionality can almost preserve their original performance.
Furthermore, we find that the embeddings after dimensionality reduction perform better than the same-dimensional embeddings produced by smaller models.
We also find that the robustness of these models to dimensionality reduction varies by task type;
while classification and clustering exhibit a more gradual performance decline, with the extent of degradation varying across models, retrieval and STS tasks tend to experience a more rapid drop in performance, with several models showing similar trends.

To investigate why such a significant reduction is feasible, we quantitatively assess the redundancy in the generated embeddings.
Specifically, we analyze the intrinsic dimensionality (ID) and isotropy of representations derived from different task-specific prompts using IsoScore~\cite{IsoScore}.
Our findings reveal that prompt-based text embedding models produce distinct representation properties depending on the task prompt.
For classification and clustering tasks, we observe lower intrinsic dimensionality and less isotropic distributions, which correlates with their high redundancy and robustness to even drastic dimensionality reduction.
In contrast, for retrieval and STS tasks---where fine-grained similarity is critical---embeddings exhibit higher intrinsic dimensionality and more isotropic distributions in the embedding space.
Moreover, we find a relationship between measures such as ID and IsoScore and the robustness of embeddings to dimensionality reduction;
embeddings for classification and clustering tasks are highly redundant and remain robust even under drastic reduction, whereas those for retrieval and STS tasks are less redundant and degrade more significantly when dimensions are reduced.

\section{Robustness of Text Embeddings for Dimensionality Reduction}
\label{sec:reduction}

In this section, we demonstrate that prompt-based text embeddings exhibit high robustness to dimensionality reduction in specific tasks and reveal that their embeddings contain redundancy.

\subsection{Evaluation Tasks}

To conduct a comprehensive analysis across various tasks, we evaluate text embedding models using the Massive Text Embedding Benchmark (MTEB)~\cite{MTEB}.
In this study, we use several English datasets from four categories: classification, clustering, retrieval,  Semantic Textual Similarity (STS).

\paragraph{Classification}
Classification tasks evaluate the quality of text embeddings by training a logistic regression classifier to predict the labels of given texts based on their corresponding embeddings.
The logistic regression classifier is trained on the training set and evaluated on the test set.
We use the default settings to train the logistic regression classifier without modifications.
Since the evaluation metrics vary by task, we adopt the default metrics.
In this study, we employ five tasks:
AmazonCounterfactualClassification~\cite{AmazonCF},
AmazonPolarityClassification~\cite{AmazonPolarity},
AmazonReviewsClassification~\cite{AmazonReview},
ImdbClassification~\cite{ImdbClassification},
and ToxicConversationsClassification.\footlink{https://kaggle.com/competitions/jigsaw-unintended-bias-in-toxicity-classification}

\paragraph{Clustering}
Clustering tasks evaluate how well the clusters formed based on distances in the embedding space align with the ground-truth clusters.
For evaluation, we use the V-Measure metric~\cite{V-Measure}, which is the default evaluation metric in MTEB.
In this study, we use three tasks;
RedditClustering, StackExchangeClustering~\cite{RedditClustering},
and ArxivClusteringS2S.\footlink{https://www.kaggle.com/datasets/Cornell-University/arxiv}

\paragraph{Retrieval\protect\footnote{
For retrieval tasks, encoding the entire document collection, which comprises millions of examples, into embeddings for each experiment is computationally infeasible.
Therefore, we use the down-sampled version officially provided by MTEB for these evaluations.
This version includes only the document sets corresponding to 250 hard negatives collected for each search query using BM25 or mE5~\cite{mE5}, and the maximum number of examples per dataset has been reduced to 1,000
(see \url{https://github.com/embeddings-benchmark/mteb/pull/1236}).
}}
Retrieval tasks assess document retrieval performance based on embeddings.
For evaluation, search queries and a collection of documents are encoded into embeddings.
The similarity between query and document embeddings is computed, and retrieval performance is assessed by checking whether the relevant document appears among the top-ranked results.
Cosine similarity is a commonly used metric, and we use it in this study as well.
We use nDCG@10 as the evaluation metric.
For the evaluation datasets, we use
MIRACL~\cite{MIRACL},
Quora,\footlink{https://quoradata.quora.com/First-Quora-Dataset-Release-Question-Pairs}
HotpotQA~\cite{HotPotQA},
DBPedia~\cite{DBPedia},
Natural Questions~\cite{NQ},
and MS MARCO~\cite{MSMARCO}.

\paragraph{STS}
Semantic textual similarity (STS) tasks evaluate how well the semantic similarity between sentence pairs, as determined by their embeddings, correlates with human-annotated similarity scores.
For evaluation, we adopt Spearman's rank correlation coefficient, in line with previous studies~\cite{SBERT,DefSent,SimCSE}.
In this study, we utilize seven tasks: STS12--16~\cite{STS12,STS13,STS14,STS15,STS16}, STS Benchmark~\cite{STSB}, and SICK-R~\cite{SICK}.

\subsection{Experimental Models}

Prompt-based text embedding models can be categorized into two types;
instruction-based text embedding models, which use natural language instructions as prompts~\cite{InstructOR,TART,Mistral-E5,NV-Embed}, and prefix-based text embedding models, which add pre-defined task-specific prefixes to the beginning of texts~\cite{E5,mE5,nomic,GTE,BGE}.

In general, instruction-based text embedding models leverage the in-context learning capabilities of large language models (LLMs) and are often built by fine-tuning LLMs.
In contrast, prefix-based text embedding models are typically constructed by fine-tuning smaller models, such as BERT~\cite{BERT}, using large-scale contrastive learning.
We include both types of models in our experiments.
For example, the instruction-based models consist of
gte-Qwen2 with 7.6B parameters and an embedding dimension of 3,584,
E5-mistral~\cite{Mistral-E5} and SFR-Embedding-2\_R\protect\footlink{https://huggingface.co/Salesforce/SFR-Embedding-2_R} each with 7.1B parameters and an embedding dimension of 4,096,
and mE5-large-inst~\cite{mE5} with 560M parameters and an embedding dimension of 1,024.
These models incorporate task-specific instructions to generate embeddings.
Other models used in our experiments include Unsup-SimCSE~\cite{SimCSE}, the small and large variants of E5~\cite{E5}, and Nomic~\cite{nomic}.
Unsup-SimCSE is a fine-tuned BERT-large with contrastive learning, while E5 uses two prefixes, ``\texttt{query:}'' and ``\texttt{passage:}.''
Nomic adapts to different tasks by employing different prefixes.
Specifically, the prefix ``\texttt{search\_query:}'' is used for retrieval queries,
``\texttt{search\_document:}'' for retrieval documents,
``\texttt{classification:}'' for classification tasks,
``\texttt{clustering:} for clustering tasks,
and for tasks such as STS in which the semantic content of the text is embedded, no prefix is used.
The task-specific prompts used for the instruction-based text embedding models are listed in Appendix~\ref{appendix:tasks}, and more detailed descriptions of each model are provided in Appendix~\ref{appendix:models}.

\subsection{Evaluation Method}

For each text embedding model, we iteratively reduce the dimensionality of the embeddings and evaluate the performance to observe the relationship between dimensionality reduction and performance degradation.
While several methods for dimensionality reduction, such as principal component analysis, are conceivable, this study simply reduces the dimensionality by taking the first $d \in \mathbb{Z}_{>0}$ dimensions of the output embeddings.
We do not normalize the output embeddings.

It is worth noting that, methods like matryoshka representation learning~\cite{MRL} exist to enable dimensionality reduction by simply taking the first $d$ dimensions of embeddings.
To ensure that the results obtained in this study are not attributable to such specific methods, we compared the performance when reducing dimensionality by randomly taking $d$ dimensions rather than taking the first $d$ dimensions.
The results showed no differences that would affect the observed results.
\comment{
We further conducted experiments using more sophisticated dimensionality reduction methods such as PCA; however, these did not reveal significant differences in the general trends.
Experimental results for dimensionality reduction methods other than taking the first $d$ dimensions,
including
taking the random $d$ dimensions,
PCA~\cite{PCA},
UMAP~\cite{UMAP},
and Isomap~\cite{Isomap}
are presented in Appendix~\ref{appendix:other-dim-reductions}.
}

\subsection{Experimental Results}

\begin{figure}[t]
\includegraphics[width=\linewidth]{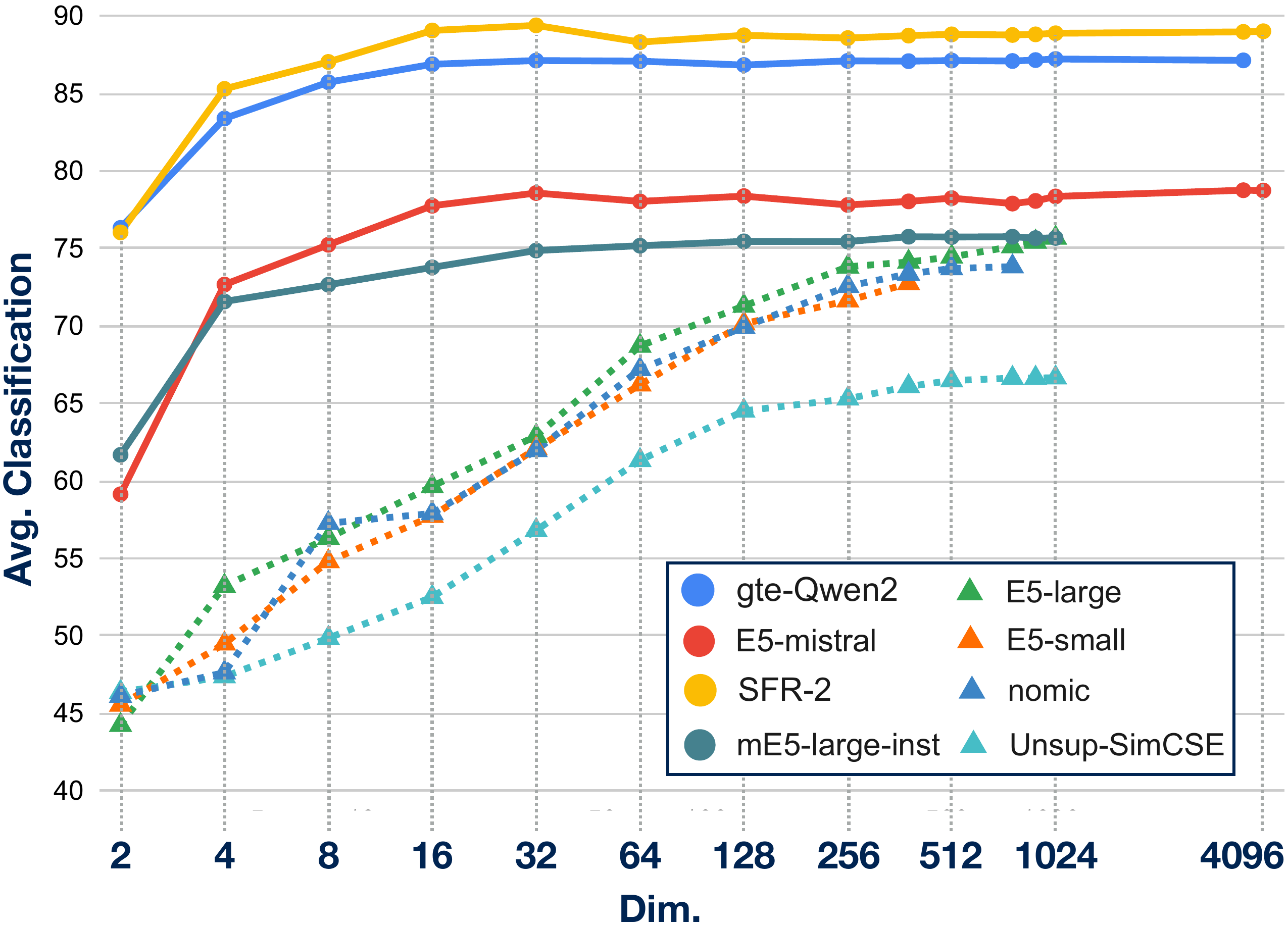}
\caption{
The relationship between the number of dimensions and the average performance on classification tasks.
The horizontal axis is logarithmic.
Circular markers with solid lines correspond to instruction-based text embedding models, whereas triangular markers with dashed lines correspond to other models.
}
\label{fig:head-classification}
\end{figure}

\begin{figure}[t]
\includegraphics[width=\linewidth]{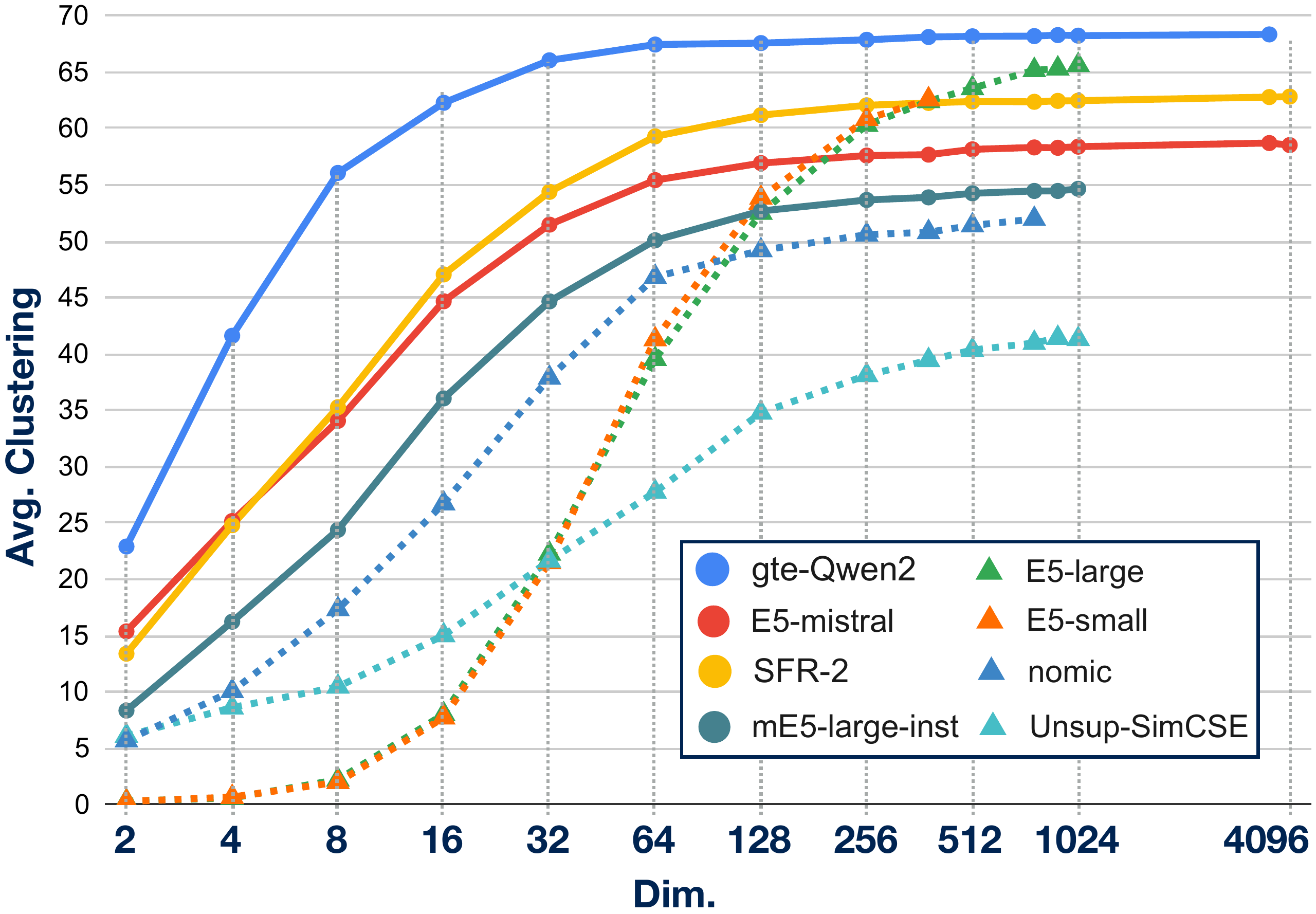}
\caption{
The relationship between the number of dimensions and the average performance on clustering tasks.
Other details are the same as in Figure~\ref{fig:head-classification}.
}
\label{fig:head-clustering}
\end{figure}

\begin{figure}[t]
\includegraphics[width=\linewidth]{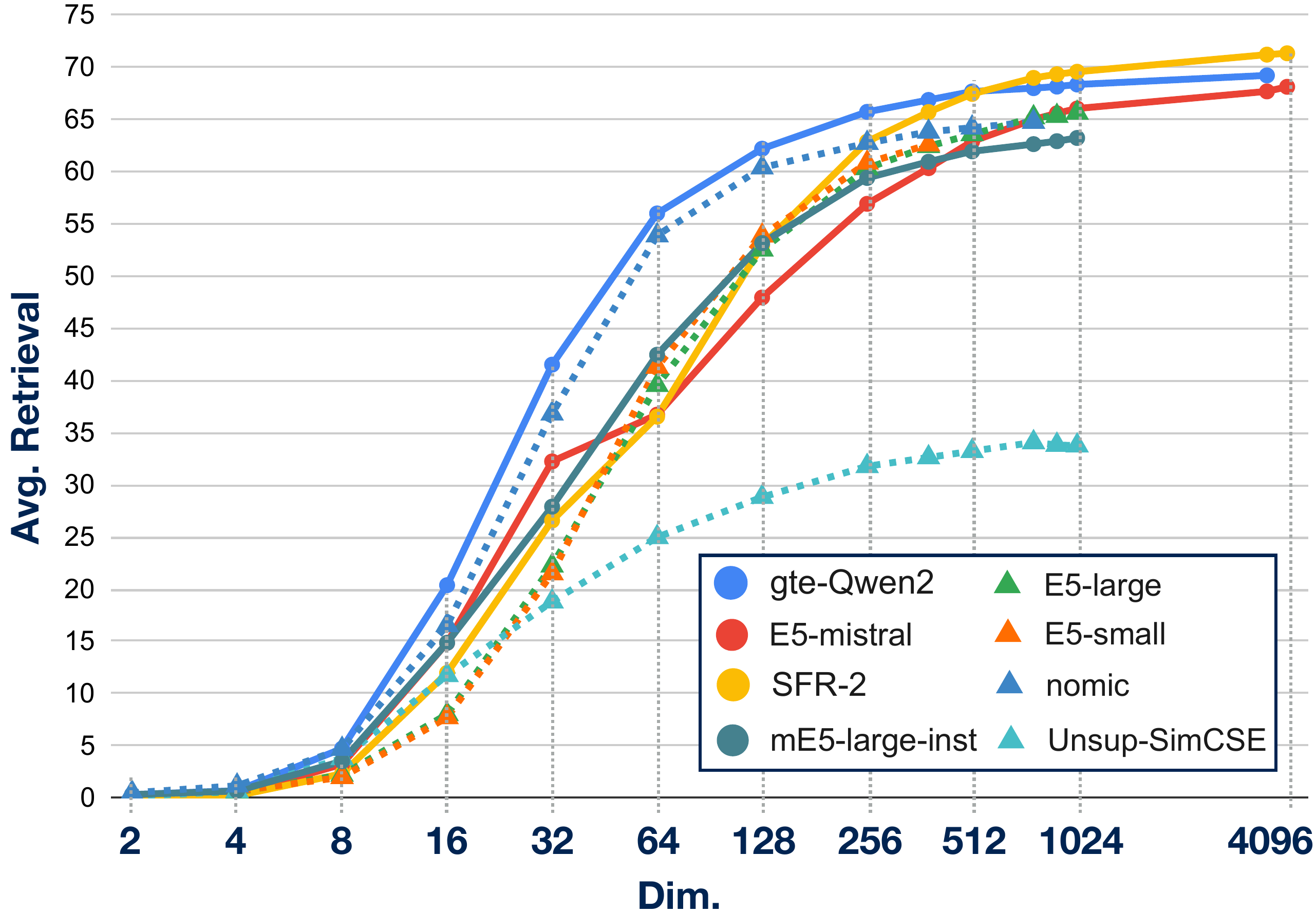}
\caption{
The relationship between the number of dimensions and the average performance on retrieval tasks.
Other details are the same as in Figure \ref{fig:head-classification}.
}
\label{fig:head-retrieval}
\end{figure}

\begin{figure}[t]
\includegraphics[width=\linewidth]{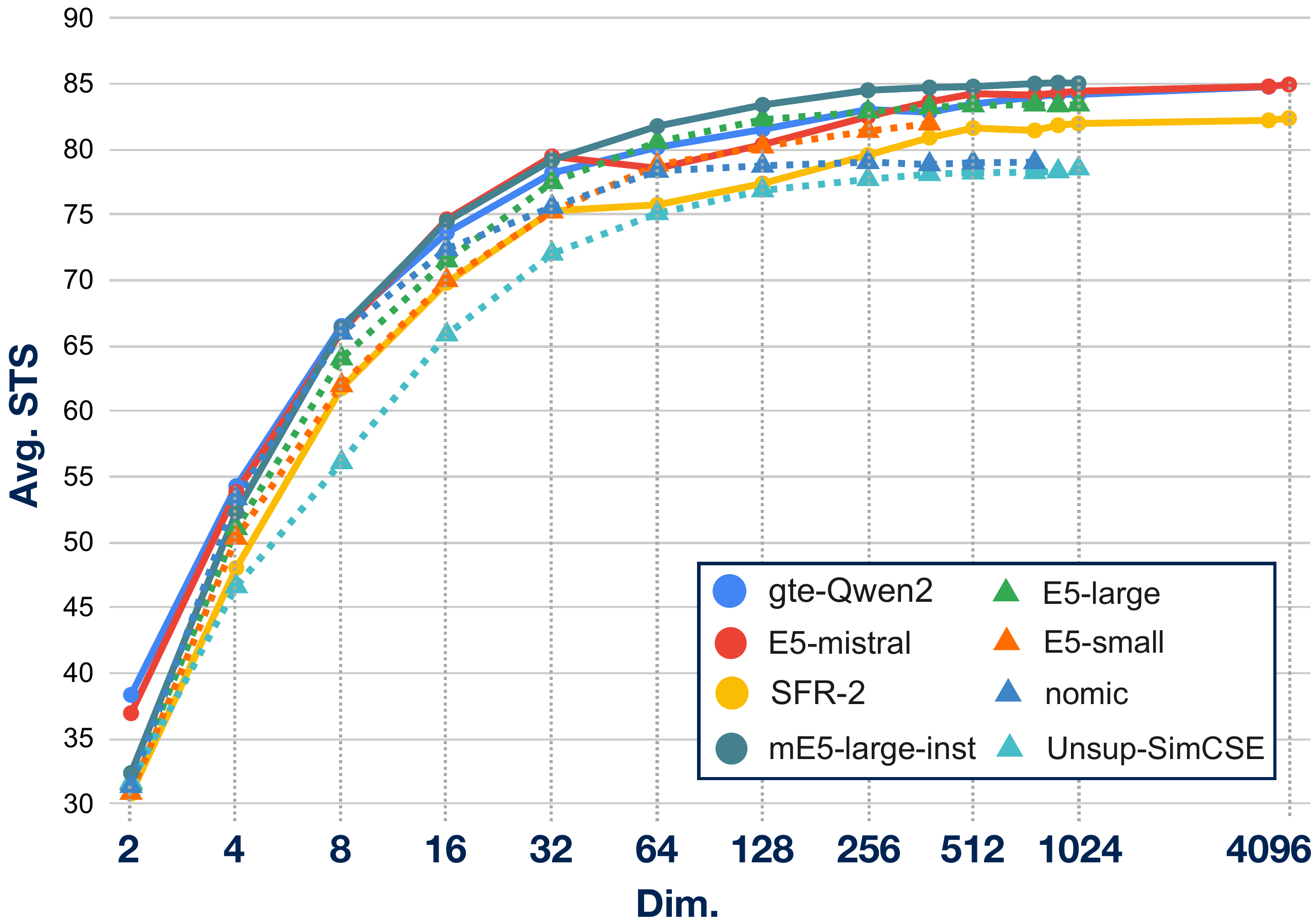}
\caption{
The relationship between the number of dimensions and the average performance on STS tasks.
Other details are the same as in Figure \ref{fig:head-classification}.
}
\label{fig:head-sts}
\end{figure}

Regarding the performance trends associated with dimensionality reduction, the results for the various tasks are presented in Figure~\ref{fig:head-classification} for classification tasks,
Figure~\ref{fig:head-clustering} for clustering tasks,
Figure~\ref{fig:head-retrieval} for retrieval tasks,
and Figure~\ref{fig:head-sts} for STS tasks.

\paragraph{Classification}
For classification tasks, we observed that the performance trends differ between instruction-based models and other models.
For instruction-based models, the degradation in performance was remarkably gradual.
In particular, both gte-Qwen2 and SFR-2 exhibited minimal performance decline when the embedding dimensionality was reduced to merely 8 dimensions (0.2\% of the original dimensions).
Notably, gte-Qwen2 achieved a score of 76.34 with just 2 dimensions, surpassing the 75.69 score obtained using the full 1024-dimensional embeddings produced by E5-large.
In contrast, models like E5-large exhibited a monotonic decrease in performance as the dimensionality was reduced.
These results suggest that instruction-based text embeddings not only generate high-quality representations for text classification, but also that the minimal performance degradation observed after dimensionality reduction implies that these embeddings exhibit significant redundancy.

\paragraph{Clustering}

The trends in clustering tasks differ slightly from those observed in classification tasks.
Although performance degradation is relatively noticeable in clustering tasks, instruction-based models remain robust to dimensionality reduction.
Specifically, we observed that LLM-based text embedding models exhibit negligible degradation even when the dimensionality is reduced to around 128 dimensions (less than 4\% of the original dimensionality).
On the other hand, in contrast to the trends observed in classification tasks, while E5 achieves high performance when using the full-dimensional embeddings, for both E5-large and E5-small the performance degradation becomes substantial.
When gte-Qwen2 embeddings are reduced to 128 dimensions (3.6\% of the original), the performance degradation is only about 0.8 points.
In contrast, when E5-large embeddings are reduced to 128 dimensions (12.5\% of the original), the performance drops by approximately 13 points, suggesting that the E5 embeddings contain little redundancy.

\paragraph{Retrieval and STS}

We observed that the performance degradation trends in retrieval and STS tasks were largely similar across all models, with performance consistently declining as the dimensionality was reduced, in contrast to the trends observed in classification and clustering tasks.
That said, when the dimensionality of gte-Qwen2 embeddings was reduced to 512 dimensions (approximately 14\% of the original), the performance loss in retrieval tasks was only about 1.5 points, remaining relatively limited.

\paragraph{Overall}

After aligning the embedding dimensionality, those produced by larger, higher-performing models consistently outperformed embeddings from smaller models.
Moreover, for classification and clustering tasks, we found that instruction-based text embeddings can maintain high performance even when only the first several dimensions of their embeddings were used, suggesting that they may be redundant.
Across tasks, classification was the most resilient to dimensionality reduction, followed by clustering, whereas retrieval and STS tasks were more sensitive.
Altogether, these results indicate that the extent of embedding redundancy varies by task.

\section{Intrinsic Dimensionality and Isotropy of Prompt-based Text Embeddings}

To investigate why the robustness to dimensionality reduction varies across tasks, we quantitatively evaluate the redundancy of the embeddings.

\subsection{Evaluation Method}

We assess the degree of redundancy in the generated text embeddings as tasks vary.
Specifically, we measure the intrinsic dimension (ID) and isotropy as indicators of redundancy, and we analyze how these metrics change as the prompt is varied across a collection of texts.

\paragraph{Intrinsic Dimension}

The intrinsic dimension refers to the number of dimensions required to capture the essential structure of data representations.
Several methods have been proposed for estimating the intrinsic dimension~\cite{Bruske,Fukunaga,MLE};
among these, we employ TwoNN~\cite{TwoNN}.
TwoNN estimates the intrinsic dimension by analyzing the ratio of distances between each point and its two nearest neighbors changes with a set of embeddings.
In high-dimensional spaces, the ratio of the distances to the first and second nearest neighbors follows a Pareto distribution for points uniformly distributed on a $d$-dimensional manifold~\cite{ID}.
TwoNN uses this property to estimate the intrinsic dimension.
Notably, TwoNN is robust even when the underlying manifold is curved or the sampling density is nonuniform, and it is computationally efficient.
We use the Python library \texttt{scikit-dimension}\footlink{https://github.com/scikit-learn-contrib/scikit-dimension} to compute the intrinsic dimensions via TwoNN.

\begin{table*}[t]
\centering
\small
\newcommand{\colw}{\hspace{5ex}}
\begin{tabular}{lc cc ccc cc cc}
\bhline
\tabH \multirow{2}{*}{Prompt Type} & \multicolumn{2}{c}{gte-Qwen2} & \multicolumn{2}{c}{E5-mistral} & \multicolumn{2}{c}{SFR-2} & \multicolumn{2}{c}{mE5-large-inst} & \multicolumn{2}{c}{nomic} \\
  &  ID & IsoScore &  ID & IsoScore &  ID & IsoScore & ID & IsoScore & ID & IsoScore \\
\bhline
\tabH Classification & 22.02 & .0052 & 22.26 & .0057 & 37.03 & .0077 & 21.85 & .0191 & 27.75 & .1556 \\
Clustering & 10.78 & .0058 & 13.01 & .0060 & 16.29 & .0138 & 17.29 & .0405 & 26.25 & .1362 \\
\hline
\tabH Retrieval &  &  &  &  &  &  & &  &  & \\
\hspace{3ex} Query & 31.90 & .0779 & 51.36 & .0761 & 81.38 & .1117 & 36.59 & .1750 & 34.74 & .2112\\
\hspace{3ex} Passage & 35.94 & .0813 & 36.69 & .0332 & 35.07 & .0555 & 35.58 & .0752 & 33.78 & .1930\\
STS & 38.47 & .0784 &  34.07 & .0439  & 41.69 & .0533  & 34.96 & .1400  & 32.84 & .2127\\
\bhline
\end{tabular}
\caption{Intrinsic dimensions and IsoScore for models using task-specific prompts, by model and prompt type.}
\label{tab:id-isoscore-main}
\end{table*}

\begin{table}[t]
\centering
\small
\begin{tabular}{llcc}
\bhline
\tabH Model & Prompt & ID & IsoScore \\
\bhline
\tabH \multirow{2}{*}{E5-small} & query: & 41.57 & .4419 \\
 & passage: & 37.60 & .3905 \\
\multirow{2}{*}{E5-large} & query: & 42.44 & .2022 \\
 & passage: & 38.50 & .1977 \\
\hline
\tabH Unsup-SimCSE &  & 27.01 & .1611 \\
\tabH BERT (CLS) &  & 20.78 & .0186 \\
\tabH BERT (Mean) &  & 17.56 & .0973 \\
\bhline
\end{tabular}
\caption{Intrinsic dimensions and IsoScore for models without task-specific prompts, by model and prompt.}
\label{tab:id-isoscore-others}
\end{table}

\paragraph{IsoScore}

IsoScore~\cite{IsoScore} is a metric used to evaluate the isotropy of embeddings.
Isotropy refers to the extent to which embeddings are uniformly distributed across the entire embedding space without bias toward specific dimensions.
Intuitively, IsoScore is computed by calculating the variance-covariance matrix of the embedding representations, normalizing it, and then measuring its deviation from the identity matrix.
IsoScore ranges from 0 to 1, with values close to 1 indicating that the embeddings are distributed isotropically and values near 0 indicating anisotropic distribution.\footnote{
Although IsoScore*~\cite{IsoScoreStar} was introduced to stabilize IsoScore computations on small datasets and enable full differentiability, our study does not involve training new embedding models;
rather, we focus on evaluating the isotropy of embeddings produced by existing models.
Since IsoScore reliably computes stable scores when the number of data samples exceeds the embedding dimensionality---and because it remains computationally efficient without requiring additional regularization---we employed the original IsoScore in our experiments.
}

\paragraph{Evaluation Procedure}

We randomly sampled 10,000 texts from English Wikipedia and obtained embeddings for each model and prompt.\footnote{
The texts from English Wikipedia were extracted from the \texttt{<p>} tags in the HTML dump at \url{https://dumps.wikimedia.org/other/enterprise_html/}.
Some texts contain multiple sentences, while others may be shorter than a full sentence.
}
The models, instructions, and prefixes used are essentially the same as those described in Section~\ref{sec:reduction}.
Additionally, we included the BERT-large \texttt{[CLS]} embedding and the average of output contextualized word embeddings in our experiments.
It is worth noting that, for instruction-based text embedding models, different prompts are used for each task even within the same task type.
Therefore, we compute intrinsic dimensions and IsoScore for each prompt and then take the average for each task type.
In retrieval tasks, different instructions or prefixes may be used for queries and documents.
Hence, we calculate the intrinsic dimension and IsoScore separately for each.
As a result, the prompt types consist of retrieval queries, retrieval documents, STS, classification, and clustering.

\subsection{Experimental Results}

The results for the instruction-based text embeddings are shown in Table~\ref{tab:id-isoscore-main}, and the results for the other text embeddings are shown in Table~\ref{tab:id-isoscore-others}.
For all models, the intrinsic dimensions were significantly smaller than the actual dimensions of the embeddings.
Larger models exhibited lower IsoScore values, whereas smaller models demonstrated relatively high isotropy.

Focusing on Table~\ref{tab:id-isoscore-main}, instruction-based text embedding models tended to have smaller intrinsic dimensions and lower IsoScore values when prompts for classification or clustering tasks were used.
In contrast, prompts for retrieval queries, retrieval documents, or STS tasks resulted in higher intrinsic dimensions and IsoScore values. 
When comparing instruction-based models, those built on LLMs exhibited greater differences in intrinsic dimensions and IsoScore values between classification/clustering tasks and retrieval/STS tasks.
Furthermore, instruction-based text embedding models (e.g., gte-Qwen2, E5-mistral, SFR-2, mE5-large-inst) showed an average difference of more than 10 in intrinsic dimension and approximately a tenfold difference in IsoScore between embeddings generated for retrieval or STS tasks and those for classification or clustering tasks.
That is, embeddings for classification and clustering tasks are relatively anisotropic, whereas those for retrieval and STS tasks are comparatively isotropic, indicating that embeddings for classification and clustering tasks are relatively more redundant.

Table~\ref{tab:id-isoscore-others} illustrates that both text embedding models not based on prompts and prefix-based models such as E5 generally exhibited relatively high intrinsic dimension and IsoScore values.
Both Unsup-SimCSE and E5-large showed higher intrinsic dimensions and IsoScore values than the original BERT-large, which aligned with previous research indicating that contrastive learning enhanced the uniformity of embeddings~\cite{SimCSE}.
E5 consistently demonstrated high values regardless of the prefix, often exhibiting larger intrinsic dimensions than those observed in LLM-based text embeddings.
These findings suggest that E5, which employed the prefix for diverse tasks such as retrieval queries and text classification, might generate embeddings with lower redundancy in order to preserve a broader range of information.
This distinction aligns with the differing requirements of downstream tasks.
Retrieval tasks require capturing subtle semantic relationships between sentences or documents, necessitating the retention of a substantial amount of information within the embeddings.
In contrast, classification and clustering tasks require only the details relevant to specific classes.
Indeed, our observations indicate that prompt-based embedding models adapt to these task characteristics by producing embeddings with higher redundancy for classification and clustering tasks, while yielding embeddings with lower redundancy for retrieval and STS tasks.

\paragraph{ID and Isotropy with Dimensionality Reduction}

\begin{figure}[t]
\includegraphics[width=\linewidth]{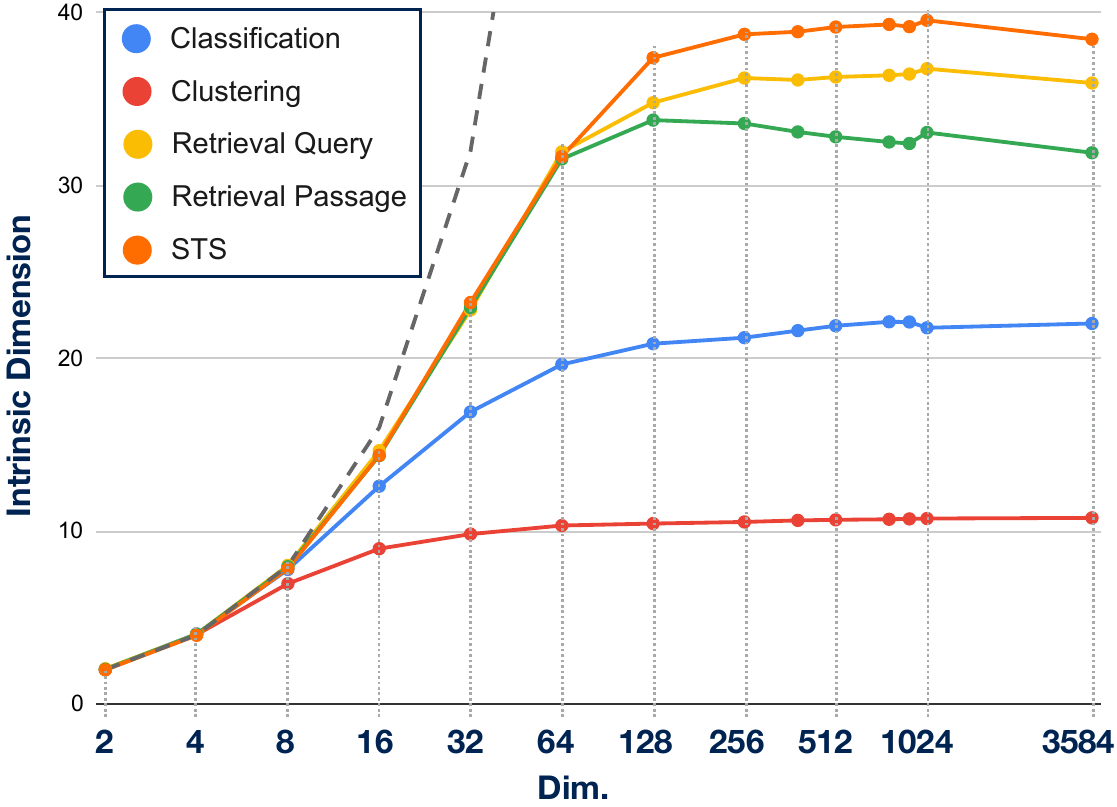}
\caption{
ID under dimensionality reduction.
The dashed line represents the actual dimensions.
}
\label{fig:reduction-id}
\end{figure}

\begin{figure}[t]
\includegraphics[width=\linewidth]{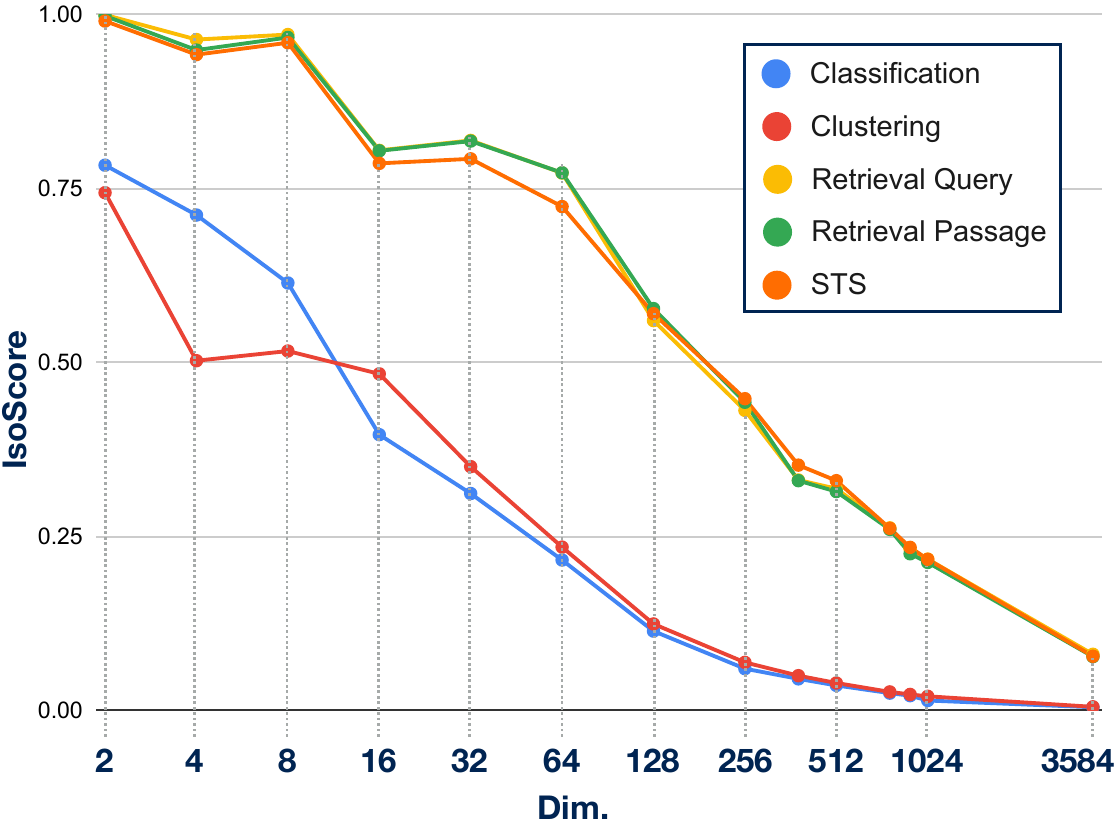}
\caption{
IsoScore under dimensionality reduction.
}
\label{fig:reduction-isoscore}
\end{figure}

As in Section~\ref{sec:reduction}, we performed dimensionality reduction on the embeddings and evaluated the changes in intrinsic dimension and isotropy.
We measured the ID and IsoScore at each dimension using embeddings from gte-Qwen2 for each prompt type, and the results are shown in Figure~\ref{fig:reduction-id} and Figure~\ref{fig:reduction-isoscore}.
Regarding ID, the ordering of the IDs for the full-dimensional embeddings did not change with dimensionality reduction;
across all prompt types, the IDs remained nearly stable until approximately 128 dimensions.
Regarding IsoScore, the trends in IsoScore differed between embeddings for classification/clustering tasks and those for retrieval/STS tasks.
Specifically, while the IsoScore for embeddings intended for classification and clustering remained around 0.75 even when reduced to 2 dimensions, the IsoScore for embeddings intended for retrieval and STS tasks nearly reached 1, indicating that the corresponding subspaces were isotropic.

\section{Related Work}

\subsection{Text Embeddings}
Early research on text embeddings focused on deriving sentence embeddings from word embeddings~\cite{SWEM,AllButTheTop,SIF,uSIF}, while later methods such as InferSent~\cite{InferSent}, Sentence-BERT~\cite{SBERT}, and Supervised SimCSE~\cite{SimCSE} fine-tuned pre-trained language models on NLI datasets~\cite{SNLI,MNLI} for richer semantic representations.
Moreover, to capture various types of information beyond semantics, prompt-based text embedding models tailor embeddings to specific tasks using prefixes~\cite{E5,mE5,nomic,GTE,BGE} or natural language instructions~\cite{InstructOR,TART,Mistral-E5,NV-Embed,GritLM}.

Prefix-based approaches like E5~\cite{E5}, multilingual E5~\cite{mE5}, GTE~\cite{GTE}, and BGE~\cite{BGE} typically fine-tune smaller models such as BERT~\cite{BERT} or XLM-RoBERTa~\cite{XLM-RoBERTa} with large-scale contrastive learning.
In contrast, instruction-based text embedding models are designed to use the target text along with a natural language instructions.
Moreover, while LLMs are originally trained using causal attention, instruction-based approaches often incorporate additional modifications to enhance contextual understanding~\cite{EchoEmbedding,PromptEOL,MetaEOL}.
Notably, one common modification is the use of bidirectional attention~\cite{LLM2Vec,NV-Embed,NV-Retriever}.
Although these models aim to generate task-specific embeddings to improve performance, whether they are truly capable of doing so and why performance improves have remained unclear.
Our study is the first to qualitatively examine what embeddings prompt-based text embedding models produce and how their properties differ across prompts.

\subsection{Embedding Dimensionality Reduction}

Several attempts to reduce the dimensionality of embeddings have long been explored, with traditional methods such as PCA~\cite{PCA} and Isomap~\cite{Isomap}.
\citet{OnTheDimensionality} investigated effective dimensionality reduction methods by training models while adjusting the pooler's output dimensionality for each reduced dimension.
Recently, learning methods designed to support post-hoc dimensionality reduction of embeddings, such as Matryoshka Representation Learning (MRL)~\cite{MRL} and Espresso Sentence Embeddings~\cite{Espresso}, have also emerged.
These methods incorporate specialized mechanisms during and after training to obtain high-performance embeddings even after dimensionality reduction.
In contrast, our study aims to demonstrate that prompt-based text embedding models can achieve high performance on certain tasks using a simple dimensionality reduction approach, without requiring additional training or specialized datasets, and to investigate the underlying factors responsible for this behavior.

\subsection{Intrinsic Dimension}

The intrinsic dimension is defined as the minimum number of dimensions required to represent the underlying structure of data representations without significant information loss, and various methods have been proposed for estimating intrinsic dimensions~\cite{Bruske,Fukunaga,MLE,TwoNN}.
Although the use of ID estimation on text embeddings is not yet widespread, there has been work applying ID-based methods to tasks such as detecting AI-generated text~\cite{AITextID}, by estimating the ID on sets of word embeddings for each text document.
\comment{
\citet{PostTempCompression} investigate the impact of the temperature parameter on model performance. They demonstrate that increasing the temperature reduces the intrinsic dimensionality and degrades retrieval performance.
To address this, they propose temperature aggregation and specialization methods, which integrate multiple temperatures directly into the contrastive training objective to balance performance and compressibility.
While \citet{PostTempCompression} consider temperature variation on a single model trained without any task-specific prompts, our findings indicate that LLM-based embedding models inherently modulate intrinsic dimensionality via instructions.
}

\subsection{Isotropy and Anisotropy}

It is well established that the contextualized word embeddings of language models are anisotropic, meaning they are predominantly distributed along a limited sub space within the embedding space.
In research on text embeddings, enhancing isotropy has been shown to improve performance on STS tasks~\cite{AllButTheTop,BERT-flow,BERT-Whitening,WhiteningBERT,ZipfianWhitening}.
In particular, training embedding models using contrastive learning techniques has been found to improve isotropy, thereby enhancing overall embedding quality~\cite{SimCSE,WhitenedCSE,OnIsotropySRL}.
Moreover, methods employing text embedding models for information retrieval~\cite{DPR} have also reported performance gains through improvements in isotropy~\cite{HIL}.

While improving the isotropy of embeddings has long been regarded as a key factor in improving their quality, recent studies have indicated that improving isotropy is not universally beneficial across all tasks.
Specifically, \citet{AnisotropyClustering} point out that enhancing isotropy does not necessarily lead to improved performance in clustering tasks, and \citet{IsotropyClustersClassifiers} argue that there exists a trade-off between the properties desirable for classification and clustering tasks, as measured by silhouette scores~\cite{Silhouettes}, and those for isotropy, which is generally preferred in STS and retrieval tasks.
These findings indicate that the optimal level of isotropy in text embeddings may vary depending on the task.
Our research supports this claim and further suggests that recent models attempt to navigate this trade-off by adjusting embeddings to exhibit varying degrees of isotropy.

\section{Conclusion and Future Work}

We demonstrated that the high-dimensional embeddings produced by prompt-based text embedding models can maintain strong performance even after dimensionality reduction by simply retaining the first several dimensions.
In particular, for classification and clustering tasks, we showed that even drastic dimensionality reduction to just a few dimensions still preserved sufficient performance.
Through analyses using intrinsic dimensionality and IsoScore, we found that prompt-based text embedding models generate embeddings with varying degrees of redundancy depending on the prompt.
Specifically, for classification and clustering tasks, embeddings exhibit lower intrinsic dimensionality and tend to be less isotropic, and that is, they have higher redundancy.
In contrast, for tasks like retrieval and STS, where fine-grained similarity is critical, embeddings tend to have higher intrinsic dimensionality and are more isotropically distributed, and that is, they have lower redundancy.

In future work, developing methods to construct embeddings with properties suitable for each task would be beneficial.  
Specifically, in contrastive learning, the temperature parameter is known to influence isotropy, with lower temperatures leading to more isotropic embeddings and higher temperatures resulting in less isotropic ones~\cite{ContrastiveLossBehaviour}.
Additionally, exploring more effective dimensionality reduction techniques for text embeddings remains an important direction.
Embeddings may contain certain crucial dimensions, and if these dimensions can be identified, it may enable more efficient dimensionality reduction.

\section*{Limitations}

In our study, we demonstrated that instruction-based text embedding models produce embeddings with different levels of redundancy depending on the prompt.
However, we have not yet clarified the underlying factors that contribute to this phenomenon.

Furthermore, we estimated intrinsic dimension and isotropy using English Wikipedia text but did not conduct a detailed analysis of how these values might vary depending on text length, domain, or differences across languages.
Expanding the range of datasets and conducting a more comprehensive analysis of downstream task performance would provide a stronger validation of how prompt-based text embeddings behave across different prompts.

\section*{Acknowledgement}

This work was partly supported by JSPS KAKENHI Grant Numbers 23KJ1134 and 24H00727.
We would also like to thank Ryo Ueda of the University of Tokyo for his insightful comments and for discussing isotropy and intrinsic dimensionality.

\bibliography{custom}

\appendix

\begin{table*}[t!]
\centering
\small
\tabcolsep 3pt
\begin{tabularx}{\textwidth}{lX}
\bhline
\tabH Task & Instruction\\
\bhline
\tabH AmazonCounterfactualClassification & Classify a given Amazon customer review text as either counterfactual or not-counterfactual\\
AmazonPolarityClassification & Classify Amazon reviews into positive or negative sentiment\\
AmazonReviewsClassification & Classify the given Amazon review into its appropriate rating category\\
ImdbClassification & Classify the sentiment expressed in the given movie review text from the IMDB dataset\\
ToxicConversationsClassification & Classify the given comments as either toxic or not toxic\\
\hline
\tabH ArxivClusteringS2S & Identify the main and secondary category of Arxiv papers based on the titles \\
RedditClustering & Identify the topic or theme of Reddit posts based on the titles \\
StackExchangeClustering & Identify the topic or theme of StackExchange posts based on the titles \\
\hline
\tabH MIRACLRetrievalHardNegatives & Given a question, retrieve Wikipedia passages that answer the question \\
QuoraRetrievalHardNegatives & Given a question, retrieve questions that are semantically equivalent to the given question \\
HotpotQAHardNegatives & Given a multi-hop question, retrieve documents that can help answer the question \\
DBPediaHardNegatives & Given a query, retrieve relevant entity descriptions from DBPedia \\
NQHardNegatives & Given a question, retrieve Wikipedia passages that answer the question \\
MSMARCOHardNegatives & Given a web search query, retrieve relevant passages that answer the query \\
\hline
\tabH STS-12 & Retrieve semantically similar text \\
STS-13 & Retrieve semantically similar text \\
STS-14 & Retrieve semantically similar text \\
STS-15 & Retrieve semantically similar text \\
STS-16 & Retrieve semantically similar text \\
STS-Benchmark & Retrieve semantically similar text \\
SICK-R & Retrieve semantically similar text \\
\bhline
\end{tabularx}
\caption{Evaluation tasks and their corresponding instructions.}
\label{tab:tasks}
\end{table*}

\begin{table*}[t]
\centering
\small
\tabcolsep 3pt
\begin{tabular}{lllcc}
\bhline
\tabH Model & HuggingFace & Prompt & Dim. & \#Params\\
\bhline
\tabH gte-Qwen2 & \hf{Alibaba-NLP/gte-Qwen2-7B-instruct} & Instruction & 3584 & 7.61B \\
E5-mistral & \hf{intfloat/E5-mistral-7b-instruct} & Instruction & 4096 & 7.11B \\
SFR-2 & \hf{Salesforce/SFR-Embedding-2\_R} & Instruction & 4096 & 7.11B \\
mE5-large-inst & \hf{intfloat/multilingual-e5-large-instruct} & Instruction & 1024 & 560M \\
\hline
\tabH nomic & \hf{nomic-ai/nomic-embed-text-v1.5} & Prefix (five types) & \hspace{1ex}768 & 137M \\
E5-small & \hf{intfloat/e5-small-v2} & Prefix (two types) & \hspace{1ex}384 & \hspace{1ex}33M \\
E5-large & \hf{intfloat/e5-large-v2} & Prefix (two types) & 1024 & 335M \\
Unsup-SimCSE & \hf{princeton-nlp/unsup-simcse-bert-large-uncased} & N/A & 1024 & 335M \\
\bhline
\end{tabular}
\caption{Details of each model.}
\label{tab:models}
\end{table*}

\section{Details of Evaluation Tasks and Prompts}
\label{appendix:tasks}
Table~\ref{tab:tasks} presents the tasks from MTEB used in our experiments, along with the instructions employed for each task when using instruction-based text embedding models.

\section{Model Details Used in Our Experiments}
\label{appendix:models}

Table~\ref{tab:models} lists the models used in the evaluation experiments.
We provide detailed descriptions of each model below.

\paragraph{Instruction-Based Text Embedding Models}

We evaluate several models that have demonstrated high performance on the MTEB.
In particular, we consider the following four models, three of which are LLM-based, while the other based on XLM-RoBERTa~\cite{XLM-RoBERTa}:
\begin{itemize}[leftmargin=1em]

\item \textbf{gte-Qwen2\footlink{https://huggingface.co/Alibaba-NLP/gte-Qwen2-7B-instruct}}:
A fine-tuned version of Qwen2 7B~\cite{Qwen2} fine-tuned for text embeddings.
This model replaces its original causal attention with bidirectional attention and is further trained on diverse multilingual datasets.
The model comprises 7.6B parameters and produces embeddings with 3,584 dimensions.

\item \textbf{E5-mistral}~\cite{Mistral-E5}:
A fine-tuned variant of Mistral 7B~\cite{Mistral} that leverages synthetic data generated by high-performant LLMs, such as GPT-4~\cite{GPT-4}.
It is a pioneering model in LLM and instruction-based text embeddings, demonstrating that the model more accurately captures the objectives of the embedding task and yields better embeddings.
The model comprises 7.1B parameters and produces embeddings with 4,096 dimensions.

\item \textbf{SFR-Embedding-2\_R (SFR-2)}\protect\footlink{https://huggingface.co/Salesforce/SFR-Embedding-2_R}:
An enhanced version of E5-mistral which is further fine-tuned using LoRA~\cite{LoRA}.
The model comprises 7.1B parameters and produces embeddings with 4,096 dimensions.

\item \textbf{mE5-large-inst}~\cite{mE5}:
A multilingual and instruction-based version of E5~\cite{E5} fine-tuned on the same datasets as E5-mistral.
Unlike the aforementioned models, mE5-large-inst is derived from XLM-RoBERTa-large~\cite{XLM-RoBERTa}.
The model comprises 560M parameters and produces embeddings with 1,024 dimensions.

\end{itemize}
Each model generates task-specific embeddings by incorporating tailored instructions into the input text.
Specifically, task instructions are prepended to the input texts prior, thereby enabling effective adaptation to a wide range of downstream tasks.
The instructions used are identical to those used in previous studies~\cite{Mistral-E5,NV-Embed}, and the ones for each task are provided in Appendix~\ref{appendix:tasks}.

\begin{table*}[ht]
    \small
    \tabcolsep 4pt
    \centering
  \begin{tabular}{lcccccccccccccc}
    \bhline
    \tabH Dim. & 2 & 4 & 8 & 16 & 32 & 64 & 128 & 256 & 384 & 512 & 768 & 896 & 1024 & 3584 \\ 
    \bhline
    \multicolumn{15}{c}{\tabH Classification}  \\
    \hline
    \tabH First    & 76.34 & 83.39 & 85.73 & 86.91 & 87.16 & 87.11 & 86.85 & 87.12 & 87.10 & 87.14 & 87.11 & 87.18 & 87.26 & 87.15 \\
    Random         & 72.87 & 81.64 & 85.42 & 86.21 & 86.79 & 86.84 & 86.79 & 87.25 & 86.98 & 87.00 & 87.15 & 87.14 & 87.16 & - \\
    PCA            & 84.86 & 85.05 & 85.22 & 85.25 & 85.31 & 85.36 & 85.40 & 85.42 & 85.43 & 85.43 & 85.44 & 85.44 & 85.44 & - \\
    UMAP           & 82.99 & 84.59 & 84.97 & 84.50 & 84.16 & 83.58 & 83.33 & 83.19 & 83.12 & 83.28 & 83.18 & 83.17 & 83.29 & - \\
    Isomap         & 83.61 & 85.19 & 85.25 & 85.30 & 85.36 & 85.45 & 85.55 & 85.66 & 85.69 & 85.72 & 85.74 & 85.75 & 85.76 & - \\
    \bhline
    \multicolumn{15}{c}{\tabH Clustering}  \\
    \hline
    \tabH First & 22.88 & 41.55 & 56.13 & 62.44 & 66.08 & 67.43 & 67.52 & 67.93 & 68.06 & 68.06 & 68.32 & 68.33 & 68.15 & 68.40 \\
    Random      & 24.70 & 41.41 & 55.33 & 62.71 & 65.89 & 67.13 & 67.44 & 67.89 & 68.12 & 68.15 & 68.25 & 68.29 & 68.26 & - \\
    PCA         & 38.64 & 55.63 & 62.93 & 66.00 & 67.81 & 68.49 & 68.40 & 68.38 & 68.43 & 68.33 & 68.37 & 68.44 & 68.48 & - \\
    UMAP        & 53.16 & 64.16 & 65.25 & 65.45 & 65.49 & 65.56 & 65.49 & 65.50 & 65.46 & 65.37 & 65.45 & 65.37 & 65.32 & - \\
    Isomap      & 42.50 & 58.68 & 63.72 & 65.51 & 66.24 & 66.11 & 66.09 & 66.00 & 66.00 & 65.82 & 65.61 & 65.82 & 65.83 & - \\
    \bhline
    \multicolumn{15}{c}{\tabH Retrieval}  \\
    \hline
    \tabH First  & \ \ 0.39  & \ \ 0.57  & \ \ 4.78   & 20.34  & 41.43  & 56.08  & 62.17  & 65.72  & 66.81  & 67.64  & 67.98  & 68.15   & 68.33  & 69.22  \\
    Random       & \ \ 0.34  & \ \ 0.88  & \ \ 5.02   & 20.02  & 40.41  & 54.98  & 61.80  & 65.52  & 66.88  & 67.43  & 68.10  & 68.31   & 68.31  & -  \\
    PCA          & \ \ 1.45  & \ \ 5.00  & 15.34  & 31.11  & 45.49  & 56.23  & 62.95  & 66.30  & 67.32  & 68.00  & 68.61  & 68.89   & 68.95  & -  \\
    UMAP         & \ \ 2.70  & \ \ 5.81  & \ \ 5.46   & \ \ 5.21   & \ \ 5.01   & \ \ 4.60   & \ \ 3.96   & \ \ 3.60   & \ \ 3.39   & \ \ 3.13   & \ \ 3.04   & \ \ 2.83    & \ \ 2.53   &  - \\
    Isomap       & \ \ 1.45  & \ \ 9.53  & 20.89  & 24.99  & 27.03  & 27.92  & 28.54  & 28.78  & 28.91  & 29.08  & 29.32  & 29.31   & 29.58  & -  \\
    \bhline
    \multicolumn{15}{c}{\tabH STS}  \\
    \hline
    \tabH First           & 38.43 & 54.29 & 66.49 & 73.65 & 78.19 & 80.15 & 81.52 & 83.04 & 82.82 & 83.44 & 84.00 & 84.18 & 84.18 & 84.76 \\
    Random          & 35.18 & 51.33 & 63.48 & 71.83 & 76.79 & 80.02 & 82.42 & 83.21 & 83.80 & 84.03 & 84.38 & 84.43 & 84.56 & - \\
    PCA             & 33.44 & 50.71 & 61.85 & 69.08 & 75.22 & 80.17 & 83.45 & 84.95 & 85.23 & 85.31 & 85.28 & 85.24 & 85.19 & - \\
    UMAP            & 53.20 & 63.40 & 65.64 & 65.38 & 65.15 & 65.54 & 65.48 & 65.71 & 65.90 & 65.47 & 65.37 & 65.38 & 65.95 & - \\
    Isomap          & 45.40 & 62.08 & 69.60 & 73.85 & 75.35 & 76.28 & 76.71 & 76.88 & 76.83 & 76.73 & 76.50 & 76.46 & 76.45 & - \\
    \bhline
  \end{tabular}
  \caption{Performance of gte-Qwen2 under various dimensionality reduction methods.}
  \label{tab:other-dim-reductions}
\end{table*}

\paragraph{Other Text Embedding Models}

Small-scale prefix-based models are highly valuable in practice, and understanding how they differ from instruction-based models is crucial; therefore, in addition to the instruction-based embedding model, we considered the following four models:
\begin{itemize}[leftmargin=1em]
    \item \textbf{Unsup-SimCSE}:
SimCSE~\cite{SimCSE} is a method for fine-tuning language models into text embedding models using contrastive learning.
In our experiments, we employ the BERT-large model fine-tuned on one million English Wikipedia sentences.\footlink{https://huggingface.co/princeton-nlp/unsup-simcse-bert-large-uncased}
This model consists of 335M parameters and outputs embeddings with 1,024 dimensions.
Notably, Unsup-SimCSE does not rely on prompts.

\item \textbf{E5 (E5-large and E5-small)}:
E5~\cite{E5} is a prefix-based text embedding model fine-tuned with large-scale contrastive learning on diverse datasets.
During contrastive learning, E5 appends prefixes such as \texttt{query:} and \texttt{passage:} to the input text, thereby enabling the effective computation of asymmetric similarities between retrieval queries and documents.
Although several model sizes are available, in this study we use the small model (E5-small)\footlink{https://huggingface.co/intfloat/e5-small-v2} comprises 33M parameters and its embedding dimension is 384 and the large model (E5-large)\footlink{https://huggingface.co/intfloat/e5-large-v2} comprises 335M parameters and its embedding dimension is 1,024.

\item \textbf{nomic}:
Nomic further develops the prefix approach used in E5 by employing five distinct prefixes tailored for different tasks.
Specifically, the prefix \texttt{search\_query:} is used for queries of retrieval,
\texttt{search\_document:} for documents of retrieval,
\texttt{classification:} for classification,
\texttt{clustering:} for clustering,
and, for tasks such as STS, where the semantic content of the text is to be embedded, it will be an empty string (i.e., no prefix).
The model comprises 137M parameters and its embedding dimension is 768.\footlink{https://huggingface.co/nomic-ai/nomic-embed-text-v1.5}
\end{itemize}

\section{Experimental Results of Other Dimensionality Reduction Methods}
\label{appendix:other-dim-reductions}

Taking the first $d$ dimensions (First) aside, we evaluate four alternative dimensionality reduction methods:
(1) Random, which selects random embedding coordinates;
(2) PCA~\cite{PCA};
(3) UMAP~\cite{UMAP};
and (4) Isomap~\cite{Isomap}.
First, the Random and PCA methods perform only linear transformations, whereas UMAP and Isomap implement nonlinear transformations.
The performance of each method are shown in Table~\ref{tab:other-dim-reductions}.
For the Random method, we fix the indices of selected dimensions prior to each task evaluation to ensure that the same subset of embedding coordinates is used consistently across runs.
Although t-SNE~\cite{t-SNE} is a well-known dimensionality reduction method, we did not employ it because it becomes computationally infeasible for projections into high-dimensional spaces: t-SNE incurs a per-iteration time complexity of $O(N^2 d)$, where $N$ denotes the number of data points.
We use gte-Qwen2 for the embedding model.
All remaining settings are identical to those specified in Section~\ref{sec:reduction}.

The results indicate that, despite its simplicity, the First method achieves performance comparable to or better than the other methods when using 8 dimensions for classification, 16 for STS, 32 for clustering, and 64 for retrieval.
Moreover, more computationally intensive methods do not necessarily yield improved results and may even degrade performance in higher-dimensional settings.

\paragraph{Classification Tasks}
Across all dimensionality reduction methods, performance decreases gradually as the number of dimensions is reduced.
At very low dimensions (2 and 4), PCA, UMAP, and Isomap outperform the other approaches;
however, for dimensions $\geq 8$, the First and Random methods consistently surpass them.
We hypothesize that PCA, UMAP, and Isomap transformations can distort the geometric structures that are critical for classification, whereas retaining the leading dimensions better preserves these intrinsic structures.

\paragraph{Clustering Tasks}
Similarly, clustering performance declines gradually for all methods.
UMAP achieves strong performance at extremely low dimensions.
PCA matches the First method across all evaluated dimensions.
The performance of the other methods deteriorates at higher dimensions.

\paragraph{Retrieval and STS Tasks}
Consistent with the observations in Section~\ref{sec:reduction}, all dimensionality reduction methods exhibit a rapid decline in retrieval performance as the number of dimensions decreases.
PCA slightly outperforms the First method in retrieval tasks; however, the improvement is marginal.
UMAP and Isomap fail to achieve competitive performance at relatively higher dimensions, likely due to distortions introduced by their nonlinear transformations.

\end{document}